\documentclass[letterpaper, 10 pt, conference]{ieeeconf}  

\IEEEoverridecommandlockouts                              
\makeatletter
\newcommand{\thickhline}{%
    \noalign {\ifnum 0=`}\fi \hrule height 1pt
    \futurelet \reserved@a \@xhline
}

\usepackage{color,soul,cite}
\usepackage{url}
\usepackage{booktabs}
\usepackage{xspace}
\usepackage{rotating}
\usepackage{graphics} 
\usepackage{cite}
\usepackage{amsmath}
\usepackage{amssymb}
\usepackage{subfig}
\usepackage{pbalance}
\usepackage{multirow}

\let\labelindent\relax
\usepackage{enumitem}
\usepackage{adjustbox}

\newcommand{\virgolette}[1]{``#1''}

\newcommand{\fref}[1]{Fig.~\ref{#1}}
\newcommand{\sref}[1]{Sec.~\ref{#1}}

\makeatletter
\let\NAT@parse\undefined
\makeatother
\usepackage[plainpages=false,hypertexnames=true,pdfnewwindow=true,backref=true,colorlinks=true,citecolor=blue,linkcolor=red,urlcolor=blue,filecolor=blue]{hyperref}

\title{\LARGE \bf
Visual Proactivity: Enhancing Human-Robot Collaboration Through Intent Communication} 

\author{Valerio Bo$^{1}$, Edison Bejarano$^{1}$, Anais Garrell$^{1}$ and Alberto Sanfeliu$^{1}$
  		\thanks{$^{1}$Institut de Robotica y Informatica Industrial (CSIC-UPC) and Universitat Politecnica de Catalunya - BarcelonaTech (UPC), Barcelona, Spain
        {\tt\small \{name.surname\}@iri.upc.edu}}
\thanks{This work was supported by 
JST Moonshot R \& D grant number JPMJMS2011 and the European project TORNADO with grant number HORIZON-CL4-2024-DIGITAL-EMERGING-01-101189557  
}   
\thanks{\copyright\ 2026 IEEE. Personal use of this material is permitted. Permission from IEEE must be obtained for all other uses, in any current or future media, including reprinting/republishing this material for advertising or promotional purposes, creating new collective works, for resale or redistribution to servers or lists, or reuse of any copyrighted component of this work in other works.}
}

\begin{document}

\maketitle
\pagestyle{empty}

\begin{abstract}
As robots transition from performing repetitive tasks to collaborating with humans, understanding human intent becomes crucial to effective interaction. Anticipation enables robots to predict human actions, while proactivity allows them to take initiative and guide human behavior toward optimal outcomes. Although research has largely focused on how robots infer and respond to human intentions, less attention has been paid to how robots communicate their own intent. 
This paper introduces \textit{visual proactivity}, a novel, simple yet effective approach that enables robots to communicate their intentions through visual feedback, influencing human behavior and enhancing transparency and fluency. We develop and evaluate proactive robotic behaviors in a human-to-robot handover scenario, where a user study validates human perception of reactive, anticipatory, and proactive behaviors. The results demonstrate that effective visual proactivity fosters better alignment and coordination, paving the way for more intuitive human-robot collaboration.

\end{abstract}

\section{Introduction}\label{sec:introduction}
The field of robotics has evolved significantly, progressing from machines designed for repetitive tasks to systems capable of complex decision-making, such as autonomous navigation \cite{floreano1996evolution} or tool selection \cite{bongard2013evolutionary}. As robots increasingly collaborate with humans, merely understanding the environment is no longer sufficient; they must also interpret human intentions, which are often unpredictable \cite{hoffman2022human}.

Understanding human intent is central to anticipatory and proactive behaviors, which are key to effective human-robot collaboration \cite{hamad2023concise}. Anticipation involves predicting human actions, while proactivity goes a step further by taking initiative and guiding human actions toward optimal outcomes \cite{rozo2016learning}. Despite their differences, these terms are sometimes used interchangeably \cite{grant2008dynamics,buyukgoz2021exploring}.

Most research on robotic proactivity has focused on recognizing and responding to human intentions through verbal and non-verbal cues \cite{sirithunge2019proactive}. Verbal cues include explicit instructions or feedback, while non-verbal cues, such as gestures, gaze, and posture, help infer human goals. However, much less attention has been paid to the inverse problem: How humans perceive and interpret robot intent. In particular, the potential of visual cues not just to clarify but to shape human decision-making remains largely unexplored. A robot that proactively communicates its intended actions can promote transparency, improve predictability, and foster trust. Yet, current approaches often overlook this dimension, focusing narrowly on intention expression for interaction clarity rather than intention guidance.

This challenge is particularly evident in collaborative assembly tasks, where seamless coordination between humans and robots is essential for efficiency and accuracy. In such settings, a robot that visually signals its next intended action can actively influence human decision-making and guide their responses, rather than simply informing. By shaping human expectations and prompting anticipatory behaviors, the robot reduces cognitive load and helps align actions, thereby minimizing disruptions. This proactive influence not only streamlines task execution but also lowers the risk of errors, ultimately fostering more intuitive, adaptive, and effective human-robot collaboration.

 \begin{figure}
	\centering
	\includegraphics[width=\columnwidth]{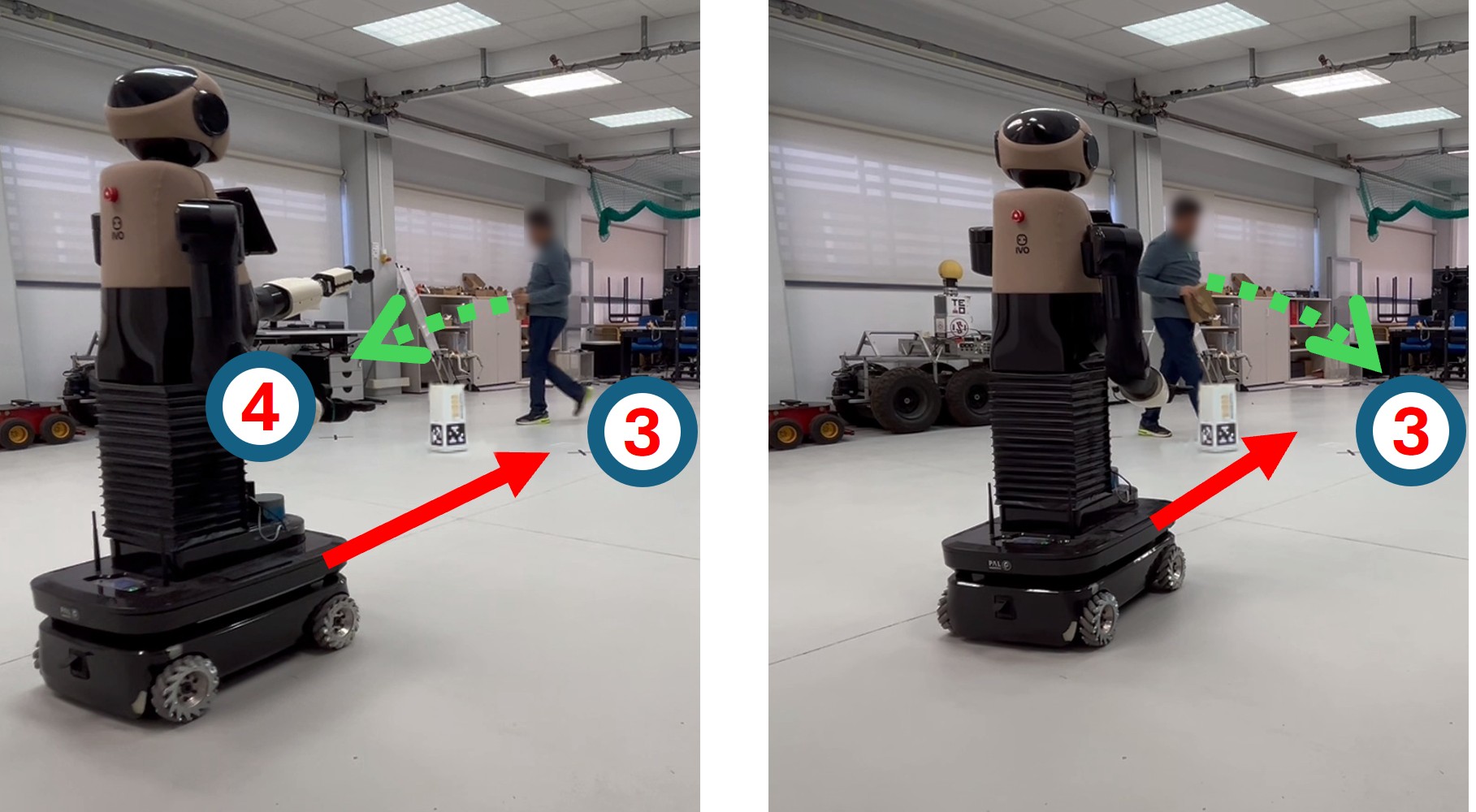}
\caption{\textbf{Visual proactive behavior.} Example in a human-to-robot handover task. The human decides to go to a specific goal (goal 4) but seeing the robot is moving towards another, the human changes his goal (goal 3).}
	\label{fig:intro}
\end{figure}

In \cite{dominguez2024anticipation}, the authors examined human perceptions of robotic behaviors, highlighting the benefits of anticipatory and proactive interactions. Building on this, we introduce \textit{visual proactivity}, a novel approach to bridging this gap (see \fref{fig:intro}). Our contributions are twofold: 
\begin{itemize}
    \item First, we developed a simple yet effective proactive robotic behavior that conveys its intentions to humans through visual feedback, aiming to influence human behavior. 
    \item Second, we demonstrate these behaviors in a robotic handover scenario, where a user study validates humans' ability to distinguish reactive, anticipatory, and proactive behaviors while assessing their impact on efficiency, predictability, and satisfaction. 
    \item Additionally, we verify that humans can accurately interpret the robot’s visual cues, fostering better alignment and coordination.
\end{itemize}
While recent social navigation research increasingly relies on real-world, multi-agent settings, our validation adopts a controlled, lab-based setup with a single human participant. This choice was intentional: it enables us to rigorously isolate the effects of visual proactivity and attribute observed behavioral changes directly to the robot's signaling strategy, without confounding factors from complex environments. Although our demonstration is framed as a handover task, the method targets the \emph{approach phase} preceding it. The same proactive signaling principle could naturally extend to subsequent phases—such as negotiating, conversing, or collaborating—making the contribution broadly applicable beyond the specific scenario tested here.  

To strengthen attribution, we also designed the evaluation around a clear baseline contrasting proactive with non-proactive behavior. This allows us to disentangle the specific impact of visual intent signaling from more general aspects of robot motion. Future work will expand this comparison to include alternative proactive strategies, such as verbal guidance, gestural cues, or multimodal signaling. These richer baselines will situate visual proactivity within the wider design space of proactive behaviors, while the current study provides the necessary first step: a clean, foundational test of its distinct contribution.  

The remainder of this paper is organized as follows: \sref{sec:proactivity} reviews related work and defines visual feedback proactivity. \sref{sec:methods} describes the methods and systems used in this study. \sref{sec:experiments} outlines the experimental setup, hypotheses, and participant distribution. \sref{sec:discussion} presents the results, and \sref{sec:conclusion} concludes the paper.

\section{Related Works}\label{sec:proactivity}
Human-robot collaboration is commonly structured around three behavioral paradigms: \textit{reactive}, \textit{anticipatory}, and \textit{proactive}. These define how robots respond to human behavior: reactive systems adapt to real-time observations, anticipatory systems leverage predictions of future human actions, and proactive systems go a step further by actively attempting to influence human decisions. While these categories are conceptually distinct, prior research often blurs the boundary between anticipation and proactivity, and pays limited attention to how robots can \textit{communicate} their own intent. Our work addresses this gap through the notion of \textit{visual proactivity}, where the robot uses its motion as an implicit signal to convey its goals and encourage human alignment.

Early research in this space has primarily emphasized how robots \textit{perceive} and respond to human behavior. Rozo et al.~\cite{rozo2016learning} and Liu et al.~\cite{liu2018learning} developed controllers that adapt to human actions in collaborative tasks, enabling responsive, flexible behavior. However, in both cases, proactivity is implemented as enhanced anticipation rather than as the deliberate expression of intent. A clearer separation of these behaviors was later formalized by Domínguez-Vidal et al.~\cite{dominguez2024anticipation}, who distinguish proactive systems as those that not only predict human actions but also influence them—an idea that forms the conceptual basis for our approach.

To achieve such influence, robots must communicate intent effectively. Much of the literature in this area relies on explicit verbal communication or augmented technologies. For instance, Andersen et al.~\cite{andersen2016projecting} and Walker et al.~\cite{walker2018communicating} employed augmented reality and visual overlays to externalize robot intentions. Sirithunge et al.~\cite{sirithunge2019proactive} reviewed non-verbal strategies, including gaze, posture, and gestures, to increase transparency. However, these methods often require additional hardware, structured environments, or are limited in expressivity. In contrast, we aim to develop an intuitive, self-contained mechanism: a robot that signals its intent solely through its physical motion, without relying on wearables.

Motion legibility plays a central role in enabling such non-verbal communication. Takayama et al.~\cite{takayama2011expressing} showed that animation-inspired principles can enhance the readability of robot actions, improving human understanding of robot goals. More recently, Schmidt-Wolf et al.~\cite{schmidt2024through} demonstrated that legibility deteriorates in complex or cluttered environments, emphasizing the need for carefully designed motion in realistic scenarios. Our work builds upon these insights by showing that legible motion can not only clarify the robot’s intent but also function as a subtle directive, guiding human partners toward shared goals.

The connection between legibility and path planning has also been explored in trajectory generation. Ngo et al.~\cite{ngo2024joint} proposed Joint Potential-Vector Fields to produce paths that are both obstacle-aware and legible to humans. While their focus is on minimizing ambiguity, their method lays the foundation for interpreting robot motion as a communicative act. We extend this idea by incorporating predictive and goal-driven cues into motion design, transforming legibility from a passive property into an active, proactive signal.

Beyond legibility, robots must also account for human spatial presence in shared tasks. Feil-Seifer et al.~\cite{feil2011people} introduced people-aware navigation, enabling robots to dynamically avoid and adapt to human movement. However, their system is largely reactive, adapting to human behavior rather than shaping it. Our approach flips this dynamic by placing the robot in a guiding role—planning motion that influences, rather than merely accommodates, human actions.

Anticipating human goals remains a critical capability for collaboration. Duarte et al.~\cite{duarte2018action} highlighted how accurate intention prediction improves fluency in joint tasks. We take this a step further: rather than simply aligning with human intentions, our robot uses predictive models to propose more efficient alternatives, leveraging motion itself to convey suggestions and encourage behavioral change.

In summary, previous research has laid essential groundwork in adaptive behavior, legibility, and intent recognition, yet the question of how robots can \textit{project} their own intent non-verbally remains underexplored. Our proposed \textit{visual proactivity} framework addresses this challenge by integrating prediction, environmental awareness, and motion design into a cohesive behavior that both reflects and communicates intent. This allows the robot to act not only as a follower or predictor but as a proactive partner—fostering intuitive and transparent collaboration without the need for verbal interaction or external devices.

\section{Methodology}\label{sec:methods}
To test our assumptions about proactivity in visual feedback, we chose a handover task as a use case. Human-robot handover is a collaborative task in which the human and the robot coordinate to transfer an object from the giver’s hand to the receiver. Although this task is typically divided into three phases (carrying, signaling, and transfer), we focused only on the first \cite{strabala2013handover}.

In this work, the human always delivers the object to one of four predefined goals. To introduce variability, three obstacles were placed between the goals, encouraging humans to take different routes. To detect and track the human body, we employed the Mediapipe library \cite{lugaresi2019mediapipe}, which operates effectively up to approximately 5 $m$. To evaluate the robot's reactive, anticipatory, and proactive behaviors, we developed several methodologies, detailed below. The reactive behavior relies on real-time human pose estimation to dynamically adjust the robot's trajectory. Anticipatory behavior enhances this by integrating motion prediction, enabling the robot to plan ahead based on expected human movements. Finally, the proactive behavior goes beyond prediction, actively influencing human decision-making through visual cues to optimize collaboration.

\begin{figure*}
	\centering
\subfloat[]{	\includegraphics[width=.6\columnwidth]{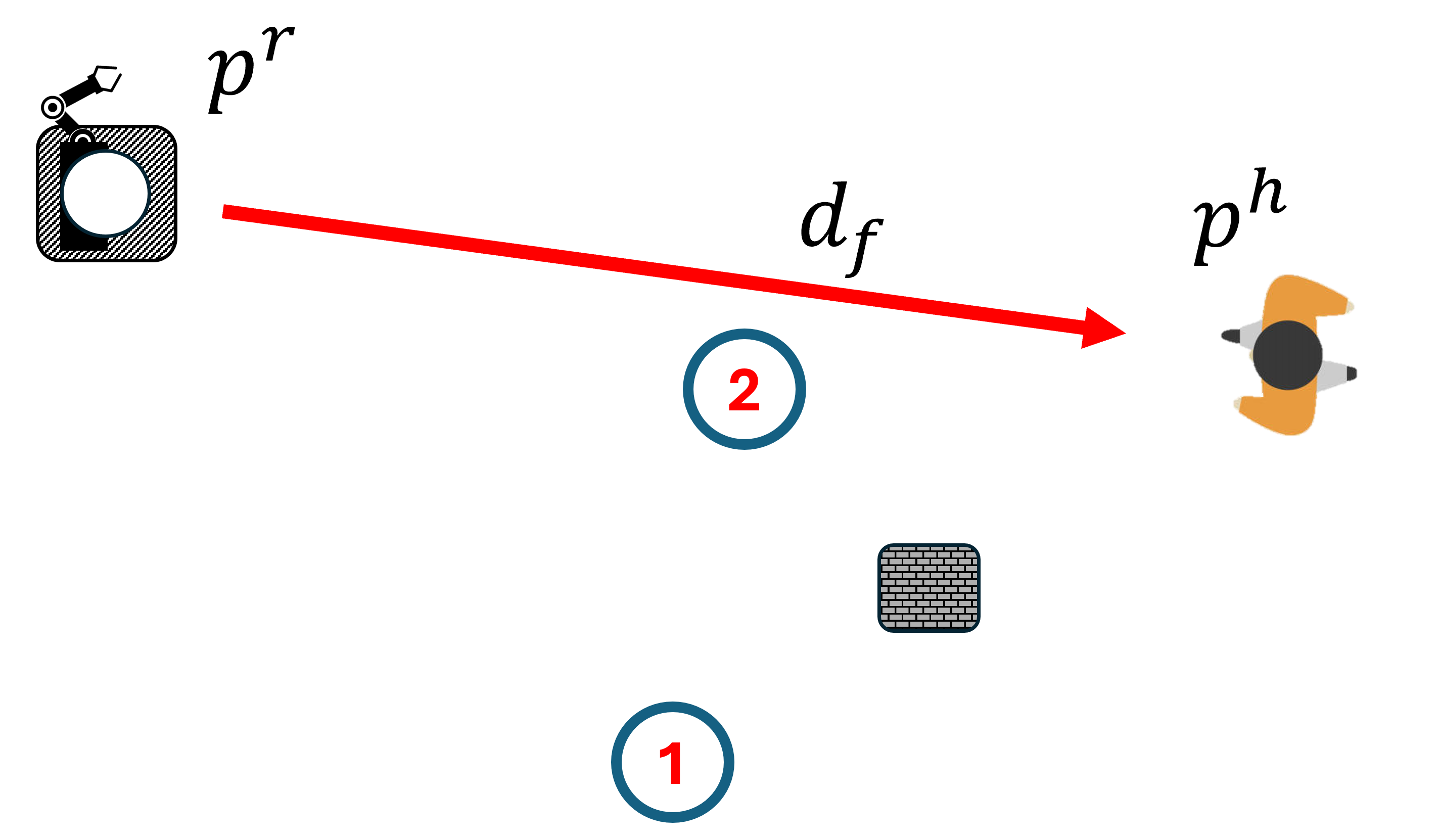} \label{fig:dir_reactive}} \hspace{0.1 pt}
\subfloat[]{	\includegraphics[width=.6\columnwidth]{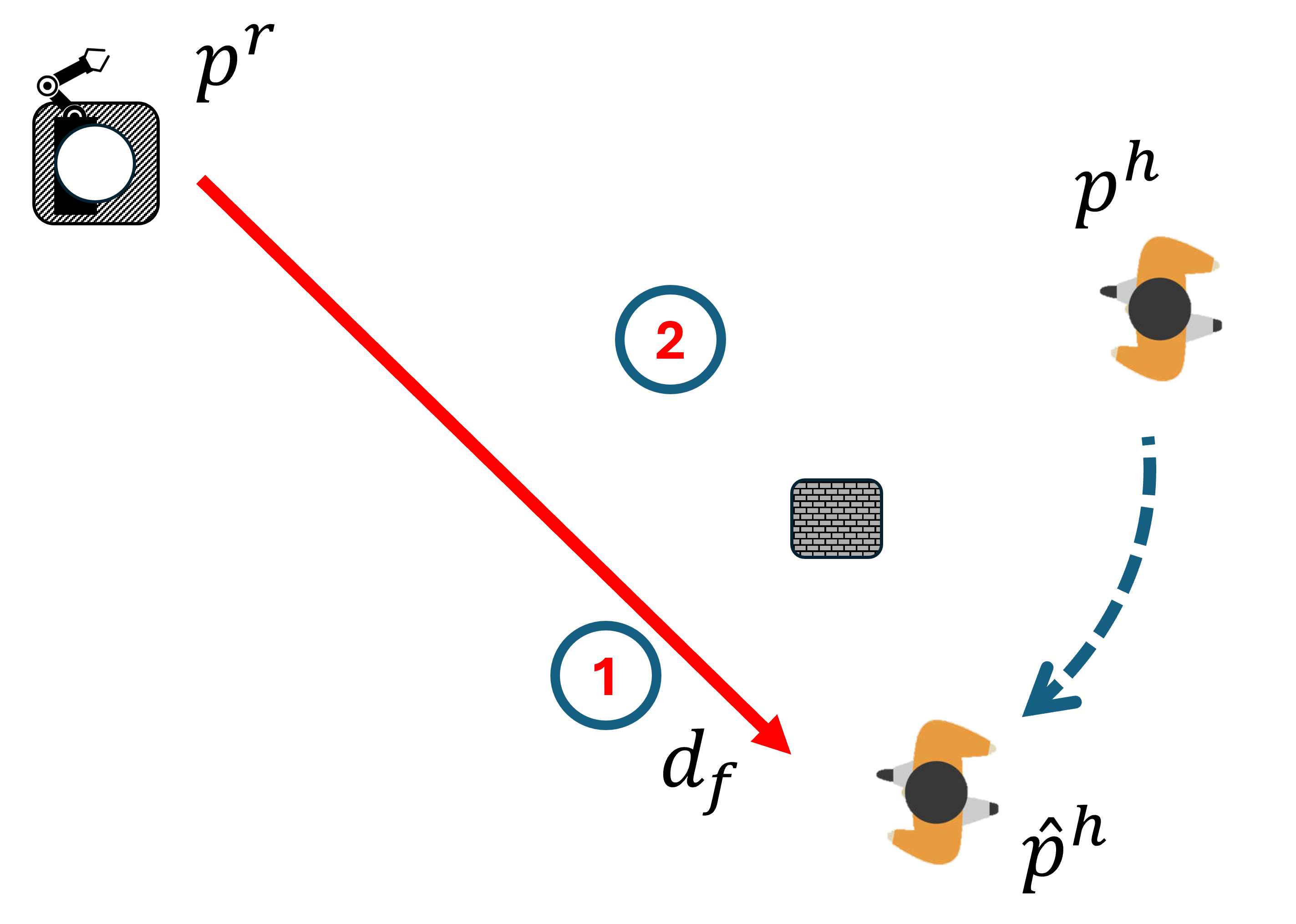} \label{fig:dir_anticipatory}}\hspace{0.1 pt}
\subfloat[]{	\includegraphics[width=.6\columnwidth]{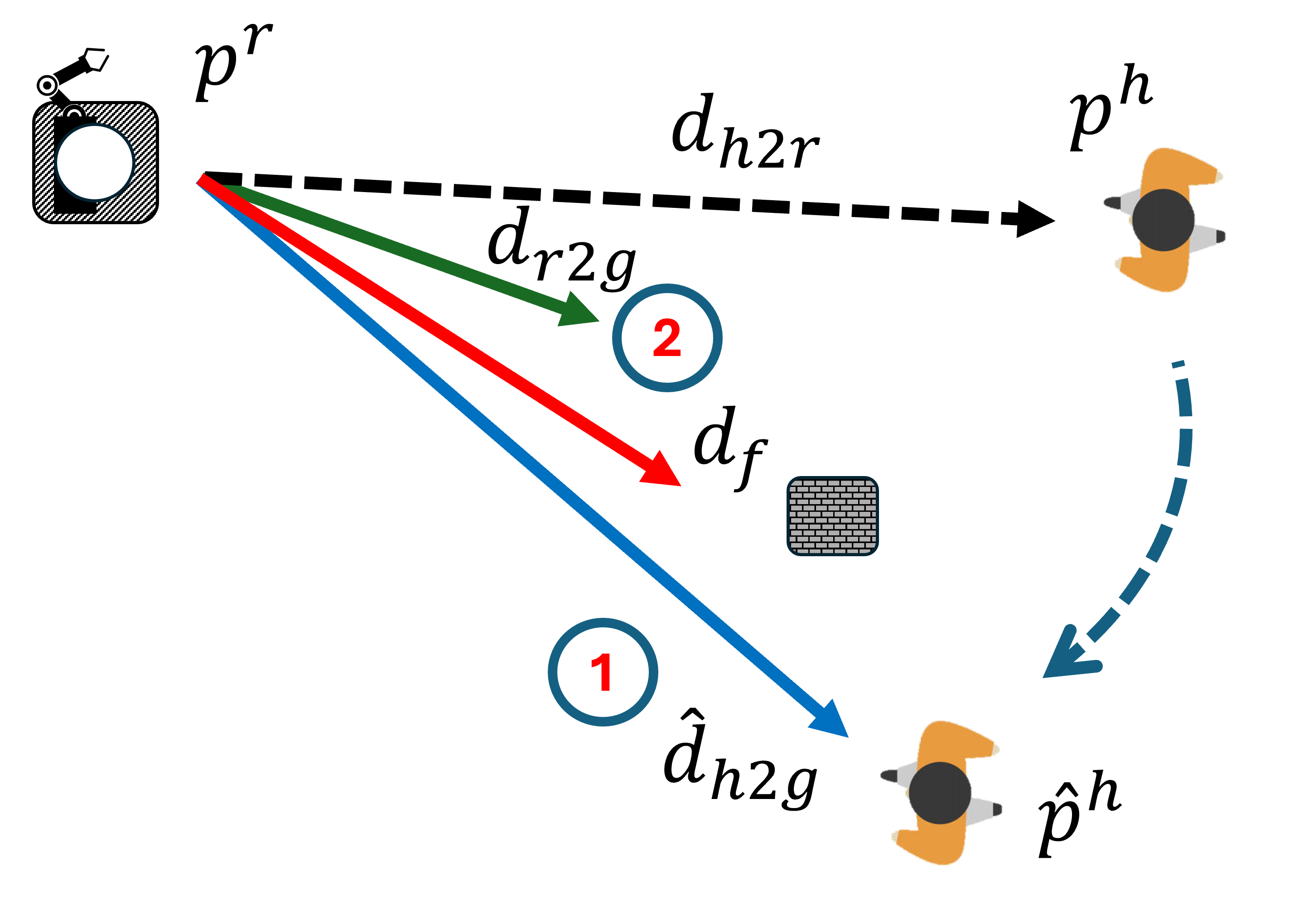} \label{fig:dir_proactive}}
\caption{\textbf{Schematic of reactive, anticipatory, and proactive approaches.} In all the figures, the approaching direction of the robot $d_f$ is represented in red. In (c), the direction between the robot and the human is depicted with dashed black, the direction between the robot and the closest goal in green, and the direction between the robot and the prediction of the human in blue. } 
	\label{fig:directions}
\end{figure*}

\subsection{Reactive Behavior Pipeline}
To implement reactive behavior, the robot was controlled using a Behavior Tree (BT). The system starts with an RGB-D camera that captures the human image, and at each time instant, the robot plans a new path based on the detected human pose. 

The reactive behavior pipeline is as follows:
\begin{enumerate}
    \item Capture an RGB image and detect human skeleton landmarks using the Mediapipe library.
    \item Extract the aligned depth image to estimate the current 3D pose of the skeleton landmarks.
    \item Plan a new path for the robot to approach the human, targeting the hip landmark with an offset of $0.2$ \textit{m} to ensure safety.
\end{enumerate}

Thus, the moving direction of the robot $d_f$ is depicted in \fref{fig:dir_reactive} and the goal coordinates are computed as:
\begin{equation*}\begin{split}
    x = p_x^r+cos(\alpha)\sqrt{{p_x^h}^2-{p_x^r}^2} \\
    y = p_y^r+sin(\alpha)\sqrt{{p_y^h}^2-{p_y^r}^2} \\
    \alpha = atan(\sqrt{{p_y^h}^2-{p_y^r}^2}/\sqrt{{p_x^h}^2-{p_x^r}^2})
\end{split}
\end{equation*}
where $p_x^h$ and $p_x^r$ represent the actual position of the human and the robot, respectively, whereas $x$, $y$, and $\alpha$ represent the goal position and yaw.
This process is repeated iteratively, with the robot continually adjusting its path to approach the human. Once the robot reaches the human or the distance threshold indicates the goal is achieved, the handover is considered complete. The robot then returns to its starting position.

\subsection{Anticipatory Behavior Pipeline}
To enable anticipatory behavior, the robot's path planning integrates predictions of human movement. A trajectory predictor \cite{laplaza2022context} estimates the human's path over the next 2.5 s based on real-time skeletal markers. This predictor employs an Attention Deep Learning model that comprises an attention module followed by a Graph Convolution Network and a Discrete Cosine Transformation, enabling more accurate, context-aware motion forecasting.

The anticipatory behavior pipeline is as follows:
\begin{enumerate}
    \item Detect the human skeleton as described in the reactive behavior.
    \item Predict the human trajectory for the next 2.5 $s$ using the trajectory predictor.
    \item Plan the robot's path based on the predicted position of the human hip landmark.
\end{enumerate}

The moving direction of the robot $d_f$ is presented in \fref{fig:dir_anticipatory}, and the goal coordinates are computed similarly to the reactive approach:
\begin{equation*}
\begin{aligned}
    x &= p_x^r + \cos(\alpha)\sqrt{{\hat{p}_x}^h{}^2 - {p_x^r}^2} \\
    y &= p_y^r + \sin(\alpha)\sqrt{{\hat{p}_y}^h{}^2 - {p_y^r}^2} \\
    \alpha &= \arctan\left(\frac{\sqrt{{\hat{p}_y}^h{}^2 - {p_y^r}^2}}{\sqrt{{\hat{p}_x}^h{}^2 - {p_x^r}^2}}\right)
\end{aligned}
\end{equation*}
where $\hat{p}_x^h$ and $p_x^r$ represent the position of the human prediction and the robot's actual position, respectively.

By continuously predicting and planning, the robot anticipates the human’s future position, enabling faster recognition of the intended delivery goal.

\subsection{Proactive Behavior Pipeline}
Developing proactive behavior is a more complex process, as the robot must go beyond merely predicting human movements. In this case, it must also integrate environmental information to suggest optimal alternative paths. To achieve this, the robot continuously evaluates both the most probable goal the human is heading toward and the closest goal relative to itself.  

In traditional approaches, when these two goals do not align, the robot may notify the user verbally and propose an alternative. However, in our method, the robot conveys the alternative path through its own movements, fully leveraging the concept of \textit{Visual Feedback Proactivity} to enhance intuitive and non-verbal communication. Thus, the objective is for the robot to navigate toward the optimal goal in a way that subtly prompts the human to reconsider their path, using motion as a form of suggestion and leveraging \textit{Visual Feedback Proactivity} to guide decision-making through intuitive, non-verbal cues.


To achieve this, the robot's goal position is determined as a weighted combination of the human's predicted goal and the shortest path. Since motion direction plays a crucial role in human-robot interactions, we evaluate the similarity between three directional vectors:
\begin{itemize}
    \item $\hat{d}_{h2g}$: the direction of the human towards the predicted goal,
    \item $d_{r2g}$: the direction of the robot towards the shortest path,
    \item $d_{h2r}$: the direction between the human and the robot.
\end{itemize}

These directions are depicted in \fref{fig:dir_proactive}. Subsequently, we compute the dot products as follows:
\begin{equation*}
\begin{aligned}
    A &= \hat{d}_{h2g} \cdot d_{h2r}, \\
    B &= d_{r2g} \cdot d_{h2r},
\end{aligned}
\end{equation*}
where $A$ and $B$ respectively measure the alignment of the human's and robot's directions to their goals relative to the direction between them.

Given these values, we introduce an additional parameter, $\gamma$, defined as a sigmoid function bounded between 0 and 1:
\begin{equation*}
\gamma = \frac{1}{1 + e^{-\frac{B}{A} - 1}},
\end{equation*}
which is used to scale the contributions of the prediction and the shortest path. The resultant direction of the robot $d_f$ is shown in \fref{fig:dir_proactive} and the goal position of the robot is then computed as:
\begin{equation*}
\begin{aligned}
    x &= \gamma p_{x_{r2g}} + (1 - \gamma) \hat{p}_{x_{h2g}}, \\
    y &= \gamma p_{y_{r2g}} + (1 - \gamma) \hat{p}_{y_{h2g}},
\end{aligned}
\end{equation*}
where $p_{r2g}$ and $\hat{p}_{h2g}$ represent the positions of the shortest goal and the predicted goal, respectively.

When $\gamma \to 0$, only the predicted goal contributes to the robot's movement. Conversely, when $\gamma \to 1$, the shortest path dominates. This enables the robot to move towards a position that balances the predicted and shortest goals, clearly communicating its preference for a different goal than the one chosen by the human.
If human motion consistently deviates from the optimal one, we also use the parameter $\gamma$ to switch to verbal proactivity rather than a globally fixed threshold. This allows the robot to notify the human that, in case its intent is not directly detected, a different goal would be optimal, thereby increasing the human's awareness of the robot's behavior. In this case, the robot would momentarily slow down, ask the human to choose a goal between the closest and the predicted, and move according to the provided answer.
While an ablation study of the threshold is outside the present scope, the consistent performance across participants suggests that its role is not overly sensitive to fine-tuning. Future work will systematically evaluate different parameterizations of $\gamma$ and explore adaptive online adjustment methods to further enhance robustness across diverse human–robot interaction scenarios.

The proactive behavior pipeline can be summarized as:
\begin{enumerate}
    \item Detect and predict the human skeleton, as described in the anticipatory behavior module.
    \item Identify the closest goal and evaluate $\gamma$.
    \item Plan the robot’s path based on the resultant position obtained through $\gamma$.
\end{enumerate}

By continuously predicting the human's goal and determining the closest possible destination, the robot can anticipate the human's future position and assess whether the current path is optimal. If the situation demands, the robot proactively suggests an alternative route through visual cues and, when appropriate, verbal communication.

\section{Experimental Setup}\label{sec:experiments}
 \begin{figure}
	\centering
	\includegraphics[width=0.95\columnwidth]{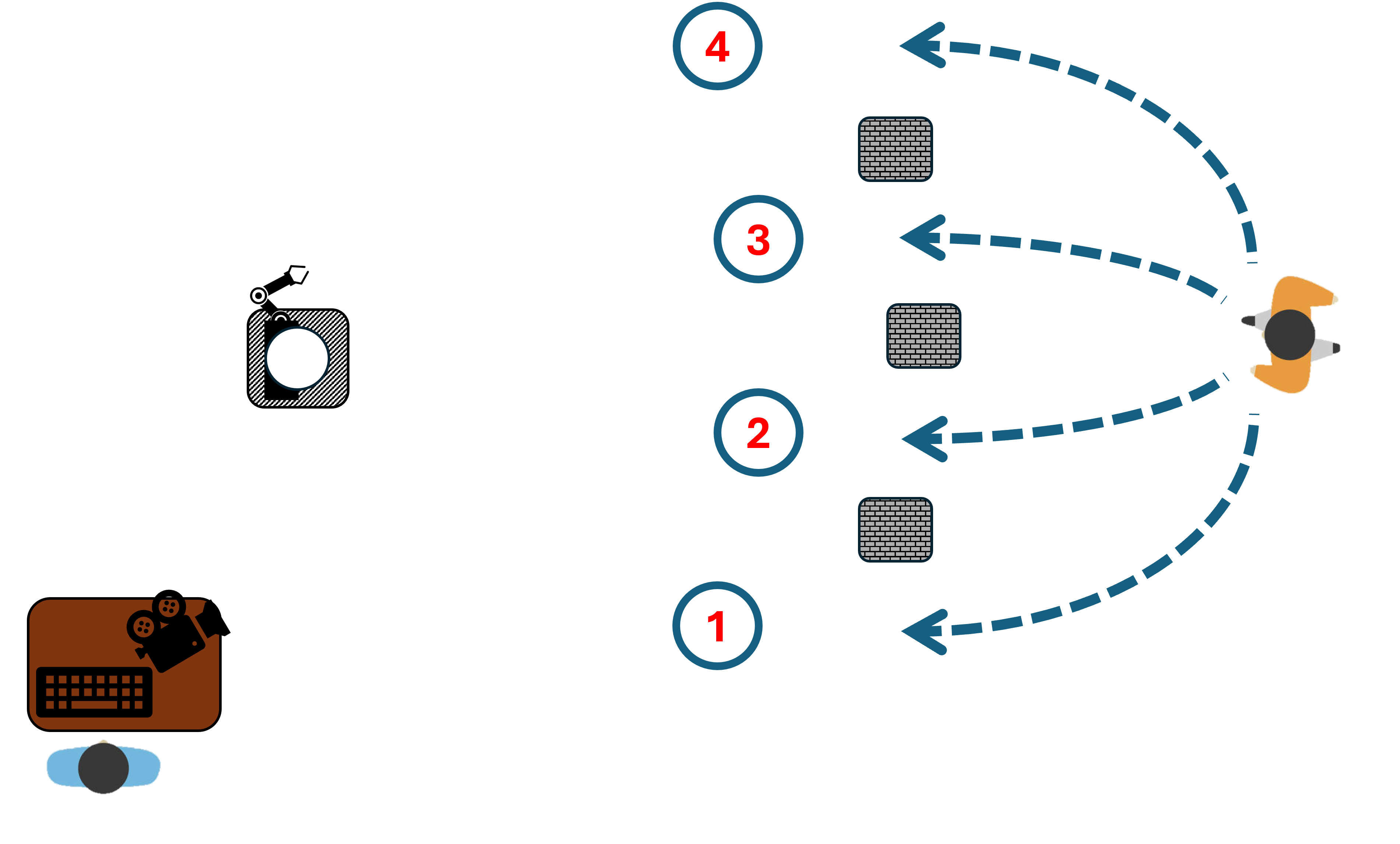}
\caption{\textbf{Experimental setup.} Four goals (numbered circles) are depicted considering the presence of three obstacles (squares). In the bottom left corner, a camera is positioned for recording. Note that volunteers were free to move as they preferred, without following precise, definite routes.}
	\label{fig:setup}
\end{figure}

\begin{figure}
	\centering
\subfloat[]{	\includegraphics[width=.9\columnwidth]{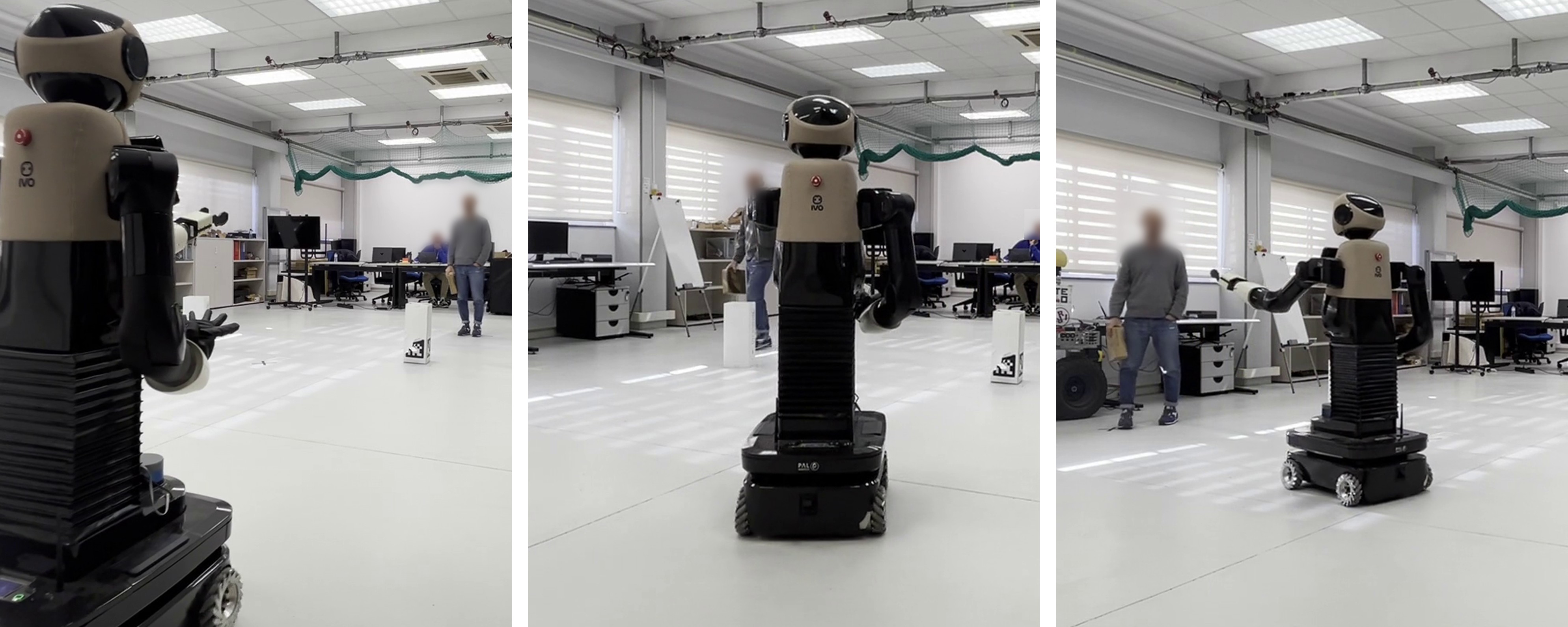} \label{fig:seq_reactive}} \\
\subfloat[]{	\includegraphics[width=.9\columnwidth]{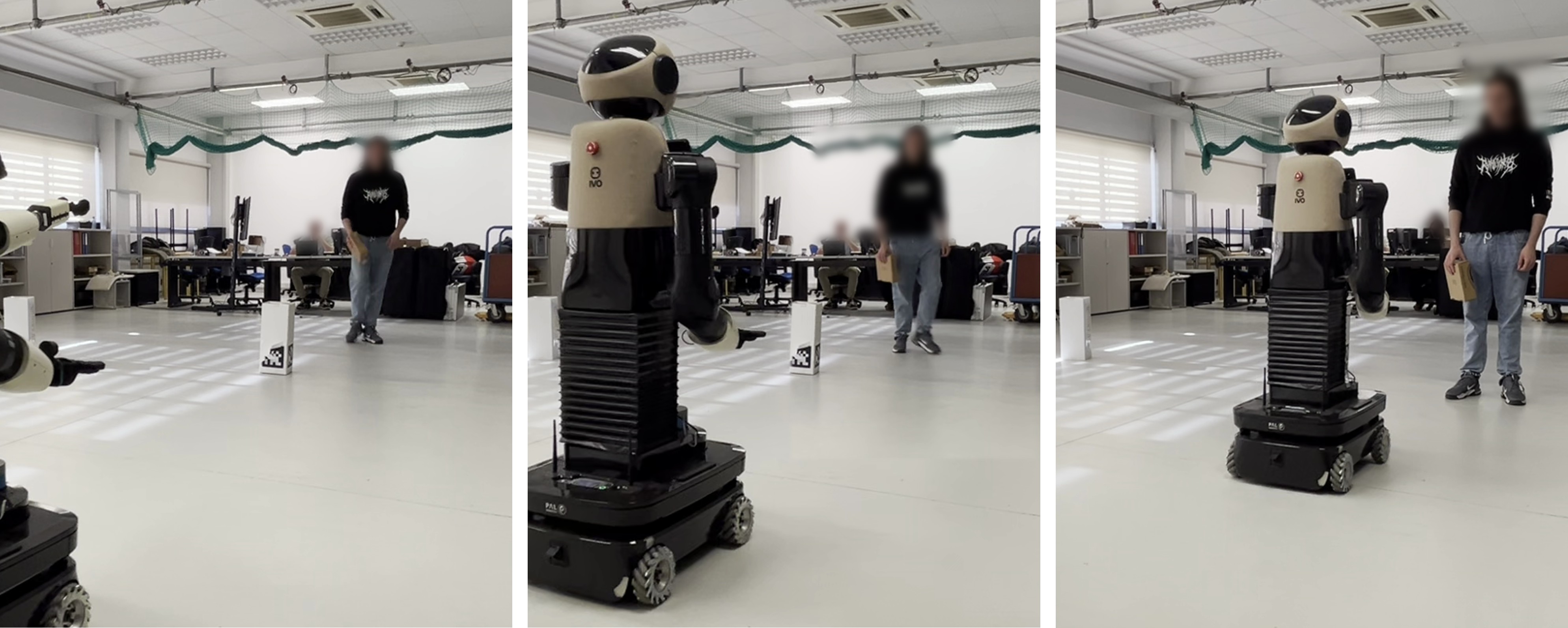} \label{fig:seq_anticipatory}}\\
\subfloat[]{	\includegraphics[width=.9\columnwidth]{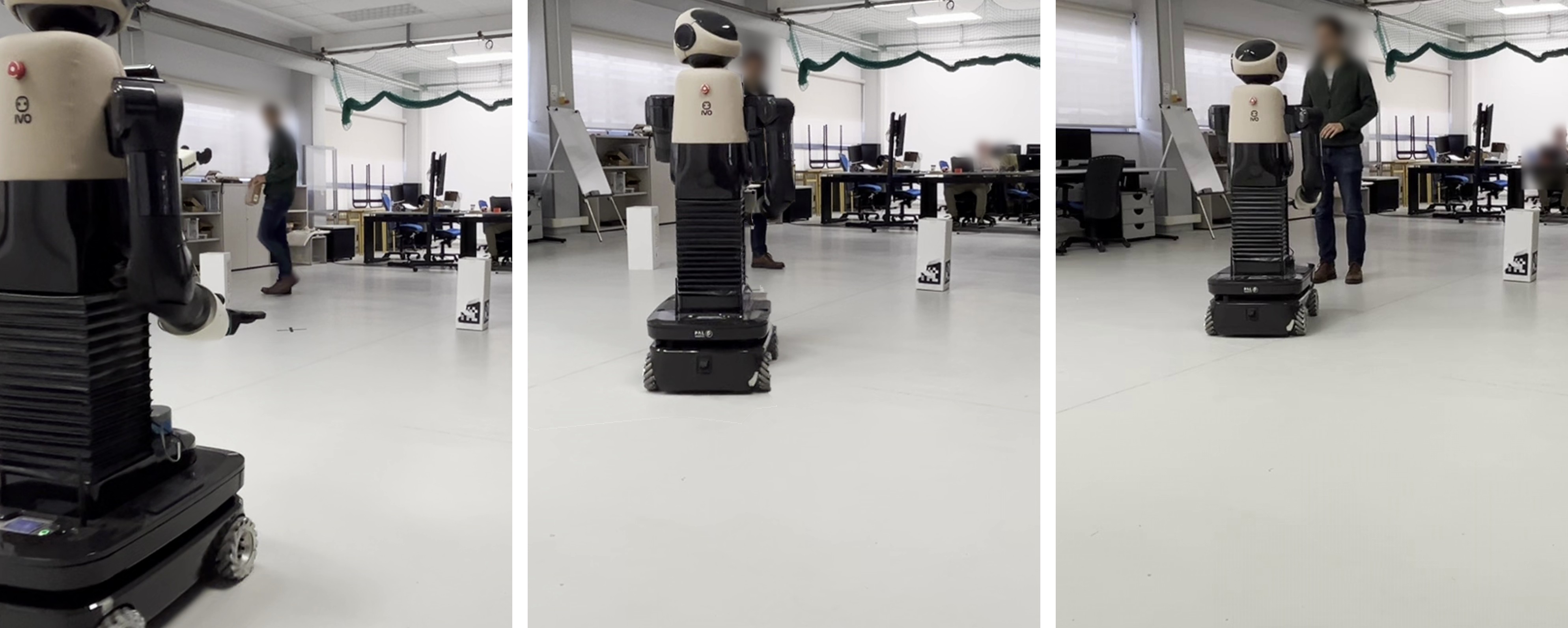} \label{fig:seq_proactive}}
\caption{\textbf{Experimental trials.} Examples of (a) reactive, (b) anticipatory, and (c) proactive approaches. In (c), the human changes his course, without the need for the robot to ask to choose between goals 3 and 4.} 
	\label{fig:sequences}
\end{figure}

In this section, we describe the real-world experiments conducted to evaluate the visual proactivity approach presented in~\sref{sec:methods}. The experimental campaign was designed to assess the following hypotheses:

\begin{itemize}
    \item \textbf{H1} -- Participants can distinguish between reactive and anticipatory robot behaviors.
    \item \textbf{H2} -- Participants can understand when the robot proposes alternative routes without relying on verbal communication.
    \item \textbf{H3} -- Anticipatory behavior improves perceived quality of human--robot interaction (HRI).
    \item \textbf{H4} -- Participants prefer proactive robot behavior over reactive navigation.
\end{itemize}

The study followed a single-blind within-subject design. Participants were informed that they would interact with a mobile robot navigating toward different goals under varying navigation strategies. However, they were not informed about the specific hypotheses of the study or the anticipatory nature of any particular behavior, to avoid biasing their perception and evaluation.
A total of 30 participants (19 male, 11 female), aged between $20$ and $64$ years ($\mu = 31.83$, $\sigma = 11.89$), were recruited. All participants provided informed consent prior to the experiments. No financial compensation was offered. The experimental protocol was approved by the Ethics Committee of the Universitat Politecnica de Catalunya (UPC), with UPC ID number 2024:020, and conducted in accordance with relevant guidelines and regulations.

\subsection{Experimental Setup}

At the beginning of each trial, both the human participant and the robot started from predefined positions, consistent across all experimental sessions, with an initial separation distance of $5$ meters. This distance was selected to ensure sufficient interaction time for trajectory adaptation and anticipatory behaviors to become clearly perceptible.

The robot used in the experiments, IVO~\cite{laplaza2022ivo}, is equipped with an omnidirectional mobile base and a humanoid torso. An RGB-D camera mounted on its head detected and tracked human skeletons, while a 2D LiDAR sensor installed on the base was employed for obstacle detection and safe navigation. The robot’s linear velocity was limited to $0.35$~m/s, corresponding to comfortable indoor shared-navigation speeds. A minimum human-robot distance of $0.2$~m was enforced to ensure safety throughout the experiments.

The scene configuration, including fixed starting positions and predefined goal locations (see \fref{fig:setup}), was chosen to ensure repeatability across participants while creating realistic path-negotiation scenarios. The spatial arrangement allowed multiple potential trajectories, thereby enabling the evaluation of different strategies under comparable conditions.

\subsection{Procedure}

Each participant performed four trials (one for each goal depicted in \fref{fig:setup}) under each of the three robot behaviors, resulting in $4 \times 3 = 12$ trials per participant. The order of both the goal locations and the robot behaviors was randomized to mitigate ordering effects. Representative trial sequences for each behavior are shown in \fref{fig:sequences}.

Before starting the recorded trials, participants completed a short familiarization trial using a neutral navigation behavior. This training phase was not included in the analysis and was introduced solely to help participants become comfortable with the robot’s motion characteristics and the environment's spatial layout.
At the beginning of the experiment, participants were instructed to walk naturally toward visually indicated goal locations while sharing the space with the robot. They were informed that the robot would also navigate toward goals in the environment, but were not given any information about the internal navigation strategies or predictive mechanisms.

After completing the four trials corresponding to a specific robot behavior, participants completed a questionnaire that included selected items from the Godspeed Questionnaire~\cite{bartneck2023godspeed} and additional validated items inspired by~\cite{dalmasso2021human}. Responses were collected using a 7-point Likert scale to evaluate perceived intelligence, safety, predictability, naturalness, and overall interaction quality.

At the end of the full experimental session, participants took part in a short semi-structured interview based on three open-ended questions: \virgolette{\textit{What do you think about the experiment in general? Have you noticed differences among the three tests? If so, which ones? What would you improve?}}
During all trials, both robot and human poses were recorded for subsequent quantitative analysis.

\section{Results and Discussion}\label{sec:discussion}
Before analyzing the questionnaire results, we removed outliers that exceeded $1.5$ times the interquartile range (IQR) from the first (Q1) and third (Q3) quartiles. This choice was made to reduce the influence of extreme responses that could disproportionately bias the non-parametric statistical tests. Importantly, the number of excluded responses was small (4 outliers, below 1\% of the total), and their removal did not qualitatively alter the distribution of the remaining data. We also conducted a power analysis to ensure we selected the appropriate sample size. Considering $p<0.05$,  \mbox{$n=30$} participants and a desired statistical power of $80\%$, we can detect effect sizes of around $\eta^2=0.123$. Given that the Shapiro-Wilk test indicated the data were not normally distributed, we employed the Friedman non-parametric test to assess significant differences across the three experimental conditions. In the following section, we will discuss each hypothesis in detail, taking into account both the questionnaire responses and the recorded human and robot trajectories. This analysis played a crucial role in confirming significant improvements in the proactive approach compared to the others.
\begin{figure}
	\centering
\subfloat[]{\includegraphics[height=4.7 cm]{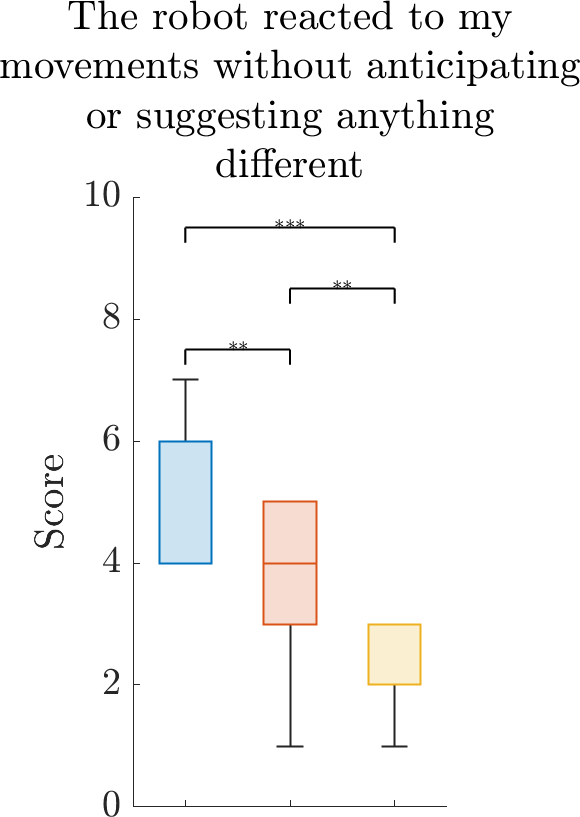} \label{fig:H1}} 
\subfloat[]{	\includegraphics[height=4.7 cm]{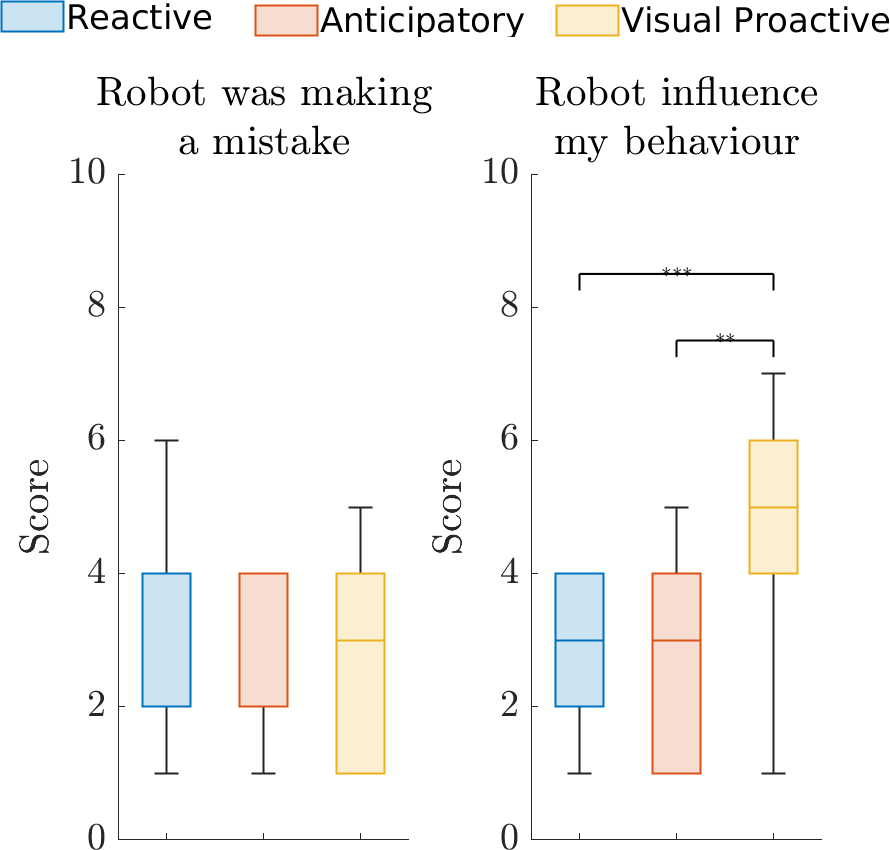} \label{fig:H2}}
\caption{\textbf{Bar chart for H1 and H2.} Results from the questions considered for (a) H1 and (b) H2. The median and interquartile range for the three behaviors are plotted. p-values are reported above the bar charts. The symbols $*$, $**$ and $***$ denote the levels of statistical significance.} 
	\label{fig:H1_H2}
\end{figure}

\begin{figure}
	\centering
\subfloat[]{\includegraphics[height=3.2 cm]{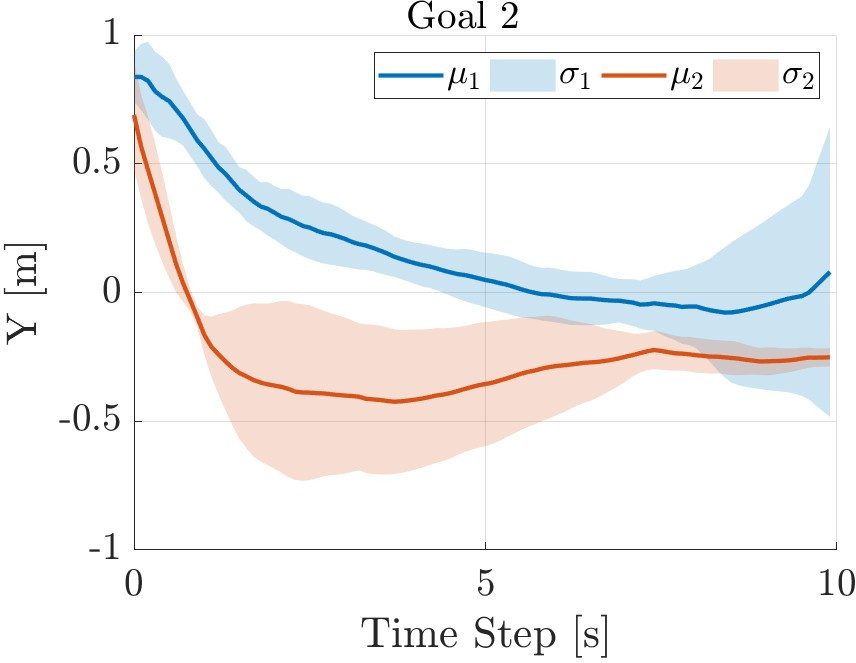} \label{fig:gmm_goal1}} 
\subfloat[]{\includegraphics[height=3.2 cm]{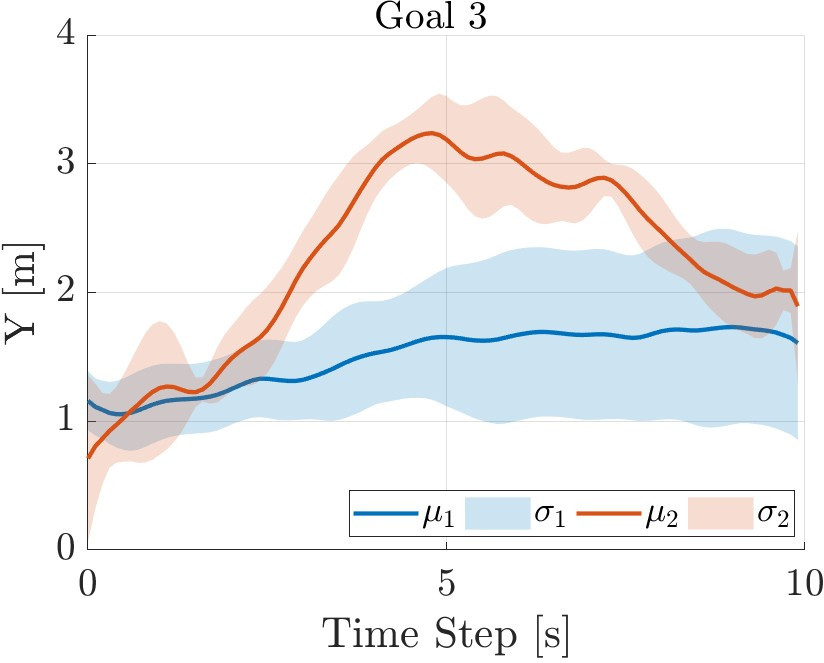} \label{fig:gmm_goal1}}
\caption{\textbf{Human trajectory clusters.} Obtained clusters from the human trajectories for Goal 2 and Goal 3 with GMM, where $\mu$ and $\sigma$ represent the mean and the standard deviation of the distribution, respectively. Two different clusters are evident where the volunteer tends to go to a different goal, but in the end, chooses the goal where the robot is moving.} 
	\label{fig:H2_gmm}
\end{figure}

\begin{figure*}
	\centering
\subfloat[]{\includegraphics[width=0.95\textwidth]{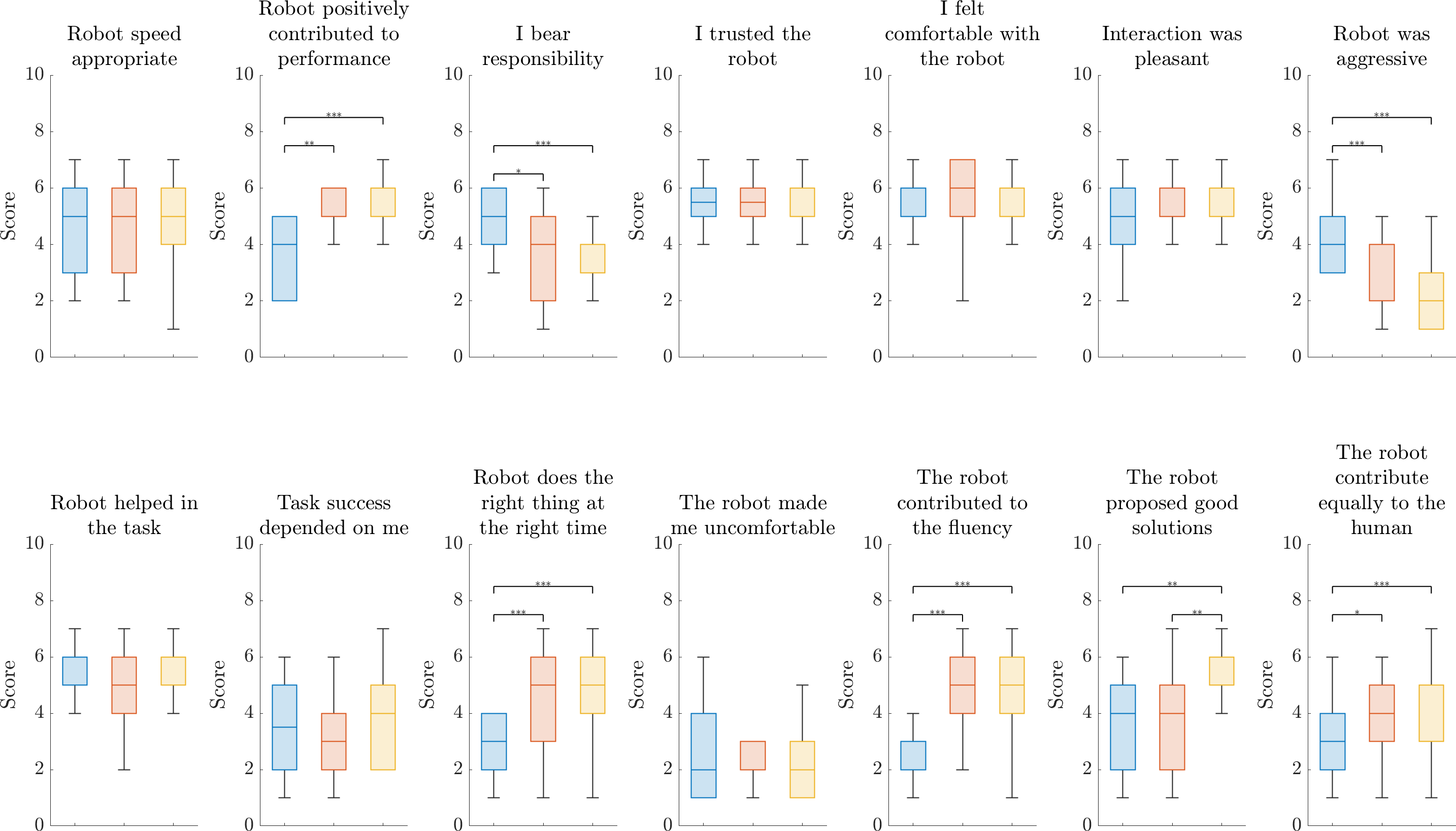} \label{fig:H3}} \\
\subfloat[]{	\includegraphics[width=0.85\textwidth]{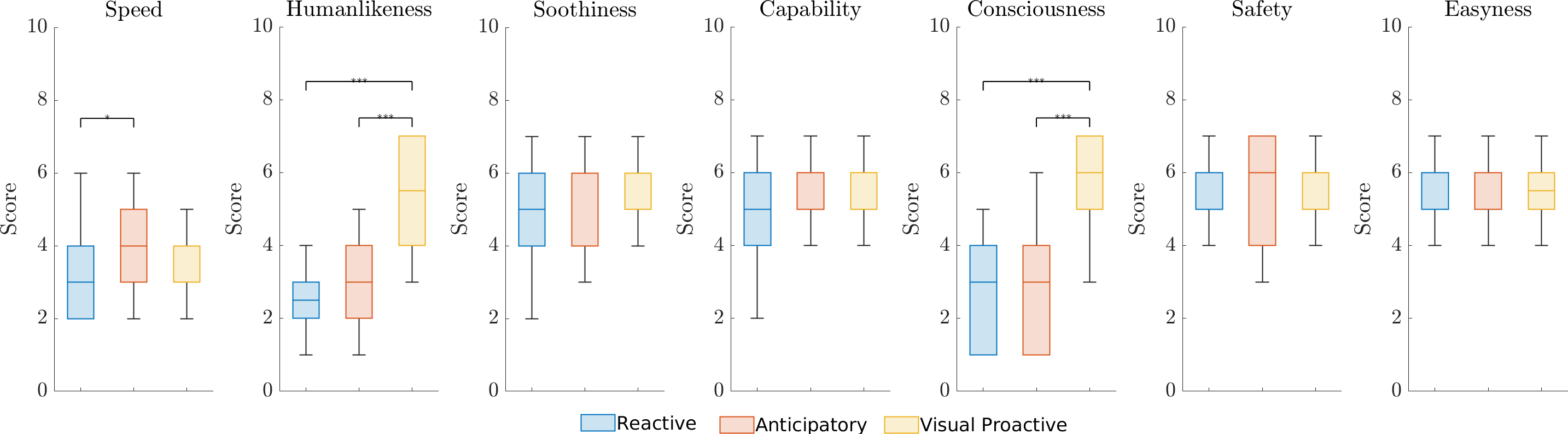} \label{fig:H4}}
\caption{\textbf{Bar chart for H3 and H4}. Results from the questions considered for (a) H3 and (b) H4. The median and interquartile range are reported for all behaviors. p-values are reported above the bar charts. } 
	\label{fig:H3_H4}
\end{figure*}


First, we tested hypothesis \textbf{H1}. This analysis used a specific question to assess whether volunteers could distinguish between reactive and anticipatory/proactive behaviors.
As shown in \fref{fig:H1}, both anticipatory and proactive behaviors have lower median scores than the reactive one (Mdn:4 and Mdn:3 vs. Mdn:6), indicating that volunteers effectively noticed differences between the behaviors. This is also confirmed by statistically significant differences ($p=0.009$ and $p<0.001$) between anticipatory and proactive behavior and reactive behavior.
Thus, \textbf{H1} is confirmed.

To evaluate \textbf{H2}, we analyzed two questions about participants' understanding of the robot's intentions. The first assessed whether the robot was perceived as moving toward an incorrect goal, and the second explored whether it influenced participants’ decision-making. As shown in \fref{fig:H2}-Left, the results are not statistically significant, and their values are similar (Mdn: $4$, $4$, $3$ for reactive, anticipatory, and proactive behaviors, respectively). This indicates that the participants perceived the proactive robot as intentionally moving toward a different goal and not making an unintended mistake. The answers to the second question show significant statistical differences between the proactive and the other behaviors ($p<0.001$ and $p=0.0014$). Thus, the analysis confirmed that the robot's movements towards a different goal were interpreted as a clear intention to influence volunteers' decision-making, as highlighted in \fref{fig:H2}-Right.

Furthermore, we analyzed participant trajectories during visual proactive behavior using Dynamic Time Warping (DTW), a technique that measures similarity between time series with temporal variations, and Gaussian Mixture Models (GMM), a probabilistic clustering method that models the underlying distribution of trajectory data as a mixture of Gaussian components. As illustrated in \fref{fig:H2_gmm}, participants initially heading towards goals 1 and 4 shifted to goals 2 and 3 (closer to the robot) without any verbal prompts. Notably, $54\%$ of trials to goals 2 or 3 fell into this cluster (red curves in \fref{fig:H2_gmm}). In $30\%$ of cases, the robot changed its course to reach the human at the predicted goal, meaning that the threshold for $\gamma$ was not reached, and for the remaining $16\%$, the robot slowed down to switch to verbal proactivity and ask the volunteer to choose between the predicted and the optimal goal. Open-ended responses corroborated this, with participants expressing understanding of the robot’s intentional movements, such as: \virgolette{\textit{During Experiment 2 (proactive), I thought the robot was moving to the wrong place at first, but I realized it was guiding me by its movements.}} These results confirm \textbf{H2}.

To verify \textbf{H3}, we analyzed 14 questions related to the robot’s perceived efficacy. As shown in \fref{fig:H3}, half of the selected results were not statistically significant (average of $p=0.4672$). However, anticipatory and especially visual proactive behaviors scored higher for \virgolette{The robot positively contributed to the performance} ($p=0.0024$ and $p<0.001$), \virgolette{The robot contributed does the right thing at the right time} (both $p<0.001$), and \virgolette{The robot contributed to the fluency} (both $p<0.001$). Moreover, although visual proactive experiments took an average of 8.3 seconds longer, this was not perceived negatively in the question \virgolette{Robot speed was appropriate} (Mdn: 5 across all behaviors). Additionally, \virgolette{The robot proposed good solutions} was rated highest for proactive behavior ($p=0.0054$ and $p=0.0042$), indicating participants found these solutions reasonable and engaging. Thus, \textbf{H3} is confirmed.

Finally, we tested \textbf{H4} using seven questions. Although most responses did not show significant differences among the three behaviors, this may be attributed to the nature of the specific areas of the experiment, such as questions related to soothing or perceived safety, which did not differ significantly across the three approaches and may have been less relevant in this context. In contrast, proactive behavior was rated considerably higher for \virgolette{Humanlikeness} and \virgolette{Consciousness} (both $p<0.001$), as shown in \fref{fig:H4}. Participants described the robot as acting more human-like and thoughtful in their open-ended responses, with comments such as: \virgolette{\textit{Experiment 3 (proactive) was more engaging, and the robot moved as one of us would have.}} Although \textbf{H4} is not fully confirmed, these findings highlight the potential of visual proactivity to enhance perceptions of humanlikeness and intentionality in robot behavior, suggesting promising directions for future research.

\section{Conclusion and Future Works}\label{sec:conclusion}
This study introduces and validates the concept of \textit{visual proactivity} as a novel approach to enhance human-robot collaboration. Through theoretical advancements and practical demonstrations in a robotic handover task, we demonstrate the effectiveness of visual feedback in improving collaboration metrics, including efficiency, predictability, and user satisfaction. The findings demonstrate that proactive robotic behaviors foster intuitive, seamless interactions by shaping human intention through visual cues. This enhances fluency and transparency in collaborative tasks.

Our experiments validate humans' ability to distinguish between reactive, anticipatory, and proactive robot behaviors. Furthermore, visual proactivity proved effective in influencing human decision-making, guiding participants to align with the robot's intended actions even in the absence of verbal communication. The observed improvements in collaboration fluency and task efficiency highlight the potential of visual feedback as a key element in the design of human-robot interaction.

Future work will focus on expanding the applicability of visual proactivity to more complex tasks and environments, such as multi-agent collaboration scenarios and dynamic, unstructured settings. Incorporating additional non-verbal modalities, such as haptic feedback and augmented reality, could further enhance the robot's ability to effectively influence human behavior. Another promising direction is exploring adaptive systems that personalize proactive behaviors based on individual user preferences, ensuring inclusivity and accessibility. Additionally, we aim to investigate how a reduced sense of responsibility may impact decision-making and ethical considerations, aligning with the principles of Responsible Research and Innovation (RRI). Finally, longitudinal studies will be conducted to evaluate the long-term impact of proactive behaviors on user trust, satisfaction, and collaboration efficiency in real-world applications.

\bibliographystyle{IEEEtran}
\bibliography{biblio}

@STRING{IROS = {Proceedings IEEE/RSJ International Conference on Intelligent Robots and Systems, IROS}}

@STRING{ICRA = {Proceedings IEEE International Conference on Robotics and Automation, ICRA}}

@STRING{ROMAN = {Proceedings IEEE International Symposium on Robot and Human Interactive Communication, ROMAN}}

@STRING{TH = {IEEE Transactions on Haptics}}

@article{floreano1996evolution,
  title={Evolution of homing navigation in a real mobile robot},
  author={Floreano, Dario and Mondada, Francesco},
  journal={IEEE Transactions on Systems, Man, and Cybernetics, Part B (Cybernetics)},
  volume={26},
  number={3},
  pages={396--407},
  year={1996},
  publisher={IEEE}
}

@article{bongard2013evolutionary,
  title={Evolutionary robotics},
  author={Bongard, Josh C},
  journal={Communications of the ACM},
  volume={56},
  number={8},
  pages={74--83},
  year={2013},
  publisher={ACM New York, NY, USA}
}

@article{hoffman2022human,
  title={Human--robot interaction challenges in the workplace.},
  author={Hoffman, Guy and Kshirsagar, Alap and Law, Matthew V},
  year={2022},
  publisher={American Psychological Association}
}

@inproceedings{dominguez2024anticipation,
  title={Anticipation and Proactivity. Unraveling Both Concepts in Human-Robot Interaction through a Handover Example},
  author={Dom{\'\i}nguez-Vidal, JE and Sanfeliu, Alberto},
  booktitle={2024 33rd IEEE International Conference on Robot and Human Interactive Communication (ROMAN)},
  pages={957--962},
  year={2024},
  organization={IEEE}
}

@article{grant2008dynamics,
  title={The dynamics of proactivity at work},
  author={Grant, Adam M and Ashford, Susan J},
  journal={Research in organizational behavior},
  volume={28},
  pages={3--34},
  year={2008},
  publisher={Elsevier}
}

@article{rozo2016learning,
  title={Learning controllers for reactive and proactive behaviors in human--robot collaboration},
  author={Rozo, Leonel and Silv{\'e}rio, Joao and Calinon, Sylvain and Caldwell, Darwin G},
  journal={Frontiers in Robotics and AI},
  volume={3},
  pages={30},
  year={2016},
  publisher={Frontiers Media SA}
}

@article{buyukgoz2021exploring,
  title={Exploring behavioral creativity of a proactive robot},
  author={Buyukgoz, Sera and Pandey, Amit Kumar and Chamoux, Marine and Chetouani, Mohamed},
  journal={Frontiers in Robotics and AI},
  volume={8},
  pages={694177},
  year={2021},
  publisher={Frontiers Media SA}
}

@inproceedings{hamad2023concise,
  title={A Concise Overview of Safety Aspects in Human-Robot Interaction},
  author={Hamad, Mazin and Nertinger, Simone and Kirschner, Robin J and Figueredo, Luis and Naceri, Abdeldjallil and Haddadin, Sami},
  booktitle={International Workshop on Human-Friendly Robotics},
  pages={1--18},
  year={2023},
  organization={Springer}
}

@article{sirithunge2019proactive,
  title={Proactive robots with the perception of nonverbal human behavior: A review},
  author={Sirithunge, Chapa and Jayasekara, AG Buddhika P and Chandima, DP},
  journal={IEEE access},
  volume={7},
  pages={77308--77327},
  year={2019},
  publisher={IEEE}
}

@article{liu2018learning,
  title={Learning proactive behavior for interactive social robots},
  author={Liu, Phoebe and Glas, Dylan F and Kanda, Takayuki and Ishiguro, Hiroshi},
  journal={Autonomous Robots},
  volume={42},
  number={5},
  pages={1067--1085},
  year={2018},
  publisher={Springer}
}

@inproceedings{laplaza2022context,
  title={Context and intention aware 3D human body motion prediction using an attention deep learning model in handover tasks},
  author={Laplaza, Javier and Moreno-Noguer, Francesc and Sanfeliu, Alberto},
  booktitle={2022 IEEE/RSJ International Conference on Intelligent Robots and Systems (IROS)},
  pages={4743--4748},
  year={2022},
  organization={IEEE}
}

@article{lugaresi2019mediapipe,
  title={Mediapipe: A framework for building perception pipelines},
  author={Lugaresi, Camillo and Tang, Jiuqiang and Nash, Hadon and McClanahan, Chris and Uboweja, Esha and Hays, Michael and Zhang, Fan and Chang, Chuo-Ling and Yong, Ming Guang and Lee, Juhyun and others},
  journal={arXiv preprint arXiv:1906.08172},
  year={2019}
}

@inproceedings{laplaza2022ivo,
  title={Ivo robot: A new social robot for human-robot collaboration},
  author={Laplaza, Javier and Rodr{\'\i}guez, Nicol{\'a}s and Dom{\'\i}nguez-Vidal, Jose Enrique and Herrero, Fernando and Hern{\'a}ndez, Sergi and L{\'o}pez, Alejandro and Sanfeliu, Alberto and Garrell, Ana{\'\i}s},
  booktitle={2022 17th ACM/IEEE International Conference on Human-Robot Interaction (HRI)},
  pages={860--864},
  year={2022},
  organization={IEEE}
}

@article{strabala2013handover,
  title={Toward seamless human-robot handovers},
  author={Strabala, Kyle and Lee, Min Kyung and Dragan, Anca and Forlizzi, Jodi and Srinivasa, Siddhartha S and Cakmak, Maya and Micelli, Vincenzo},
  journal={Journal of Human-Robot Interaction},
  volume={2},
  number={1},
  pages={112--132},
  year={2013},
  publisher={Journal of Human-Robot Interaction Steering Committee}
}

@inproceedings{andersen2016projecting,
  title={Projecting robot intentions into human environments},
  author={Andersen, Rasmus S and Madsen, Ole and Moeslund, Thomas B and Amor, Heni Ben},
  booktitle={2016 25th IEEE International Symposium on Robot and Human Interactive Communication (RO-MAN)},
  pages={294--301},
  year={2016},
  organization={IEEE}
}

@inproceedings{walker2018communicating,
  title={Communicating robot motion intent with augmented reality},
  author={Walker, Michael and Hedayati, Hooman and Lee, Jennifer and Szafir, Daniel},
  booktitle={Proceedings of the 2018 ACM/IEEE International Conference on Human-Robot Interaction},
  pages={316--324},
  year={2018}
}

@article{duarte2018action,
  title={Action anticipation: Reading the intentions of humans and robots},
  author={Duarte, Nuno Ferreira and Rakovi{\'c}, Mirko and Tasevski, Jovica and Coco, Moreno Ignazio and Billard, Aude and Santos-Victor, Jos{\'e}},
  journal={IEEE Robotics and Automation Letters},
  volume={3},
  number={4},
  pages={4132--4139},
  year={2018},
  publisher={IEEE}
}

@inproceedings{takayama2011expressing,
  title={Expressing thought: improving robot readability with animation principles},
  author={Takayama, Leila and Dooley, Doug and Ju, Wendy},
  booktitle={Proceedings of the 6th international conference on Human-robot interaction},
  pages={69--76},
  year={2011}
}

@article{schmidt2024through,
  title={Through the clutter: Exploring the impact of complex environments on the legibility of robot motion},
  author={Schmidt-Wolf, Melanie and Becker, Tyler and Oliva, Denielle and Nicolescu, Monica and Feil-Seifer, David},
  journal={arXiv preprint arXiv:2406.00119},
  year={2024}
}

@inproceedings{ngo2024joint,
  title={Joint Potential-Vector Fields for Obstacle-Aware Legible Motion Planning},
  author={Ngo, Huy Quyen and Steinfeld, Aaron},
  booktitle={2024 33rd IEEE International Conference on Robot and Human Interactive Communication (ROMAN)},
  pages={1856--1863},
  year={2024},
  organization={IEEE}
}

@inproceedings{feil2011people,
  title={People-aware navigation for goal-oriented behavior involving a human partner},
  author={Feil-Seifer, David and Matari{\'c}, Maja},
  booktitle={2011 IEEE International Conference on Development and Learning (ICDL)},
  volume={2},
  pages={1--6},
  year={2011},
  organization={IEEE}
}

@incollection{bartneck2023godspeed,
  title={Godspeed questionnaire series: Translations and usage},
  author={Bartneck, Christoph},
  booktitle={International handbook of behavioral health assessment},
  pages={1--35},
  year={2023},
  publisher={Springer}
}

@inproceedings{dalmasso2021human,
  title={Human-robot collaborative multi-agent path planning using Monte Carlo tree search and social reward sources},
  author={Dalmasso, Marc and Garrell, Ana{\'\i}s and Dom{\'\i}nguez, Jos{\'e} Enrique and Jim{\'e}nez, Pablo and Sanfeliu, Alberto},
  booktitle={2021 IEEE international conference on robotics and automation (ICRA)},
  pages={10133--10138},
  year={2021},
  organization={IEEE}
}

\end{document}